\documentclass[review]{elsarticle}

\usepackage{lineno,hyperref}
\usepackage{array}
\usepackage{amsmath}
\usepackage{amsthm}
\usepackage{amssymb}
\usepackage{multirow}
\usepackage{multicol}
\usepackage{color}
\usepackage{pifont}
\usepackage{bbm}
\usepackage{bm}
\usepackage{algorithm}
\usepackage{algorithmic} 
\usepackage{setspace}
\usepackage{xcolor}
\usepackage{booktabs}

\modulolinenumbers[5]
\journal{Journal of \LaTeX\ Templates}

\begin{document}

\begin{frontmatter}

\title{TENet: Targetness Entanglement Incorporating with Multi-Scale Pooling and Mutually-Guided Fusion for RGB-E Object Tracking}


\author[mymainaddress]{Pengcheng Shao}

\author[mymainaddress]{Tianyang Xu}

\author[mymainaddress]{Zhangyong Tang}

\author[mymainaddress]{Linze Li}

\author[mymainaddress]{Xiao-Jun Wu\corref{mycorrespondingauthor}}
\cortext[mycorrespondingauthor]{Corresponding author}
\ead{wu_xiaojun@jiangnan.edu.cn}

\author[b]{Josef Kittler}

\address[mymainaddress]{School of Artificial Intelligence and Computer Science, Jiangnan University, Wuxi 214122, China}
\address[b]{Centre for Vision, Speech and Signal Processing (CVSSP), University of Surrey, Guildford GU2 7XH, UK}

\begin{abstract}
There is currently strong interest in improving visual object tracking by augmenting the RGB modality with the output of a visual event camera that is particularly informative about the scene motion. However, existing approaches perform event feature extraction for RGB-E tracking using traditional appearance models, which have been optimised for RGB only tracking,  without adapting it for the intrinsic characteristics of the event data. To address this problem, we propose an Event backbone (Pooler), designed to obtain a high-quality feature representation that is cognisant of the innate characteristics of the event data, namely its sparsity. In particular, Multi-Scale Pooling is introduced to capture all the motion feature trends within event data through the utilisation of diverse pooling kernel sizes.The association between  the derived RGB and event representations is established by an innovative module performing adaptive Mutually Guided Fusion (MGF). Extensive experimental results show that our method significantly outperforms state-of-the-art trackers on two widely used RGB-E tracking datasets, including VisEvent and COESOT, where the precision and success rates on COESOT are improved by 4.9\% and 5.2\%, respectively. Our code will be available at https://github.com/SSSpc333/TENet.
\end{abstract}


\end{frontmatter}

\section{Introduction}

\begin{figure}
\centering
\includegraphics[width=1.0\linewidth]{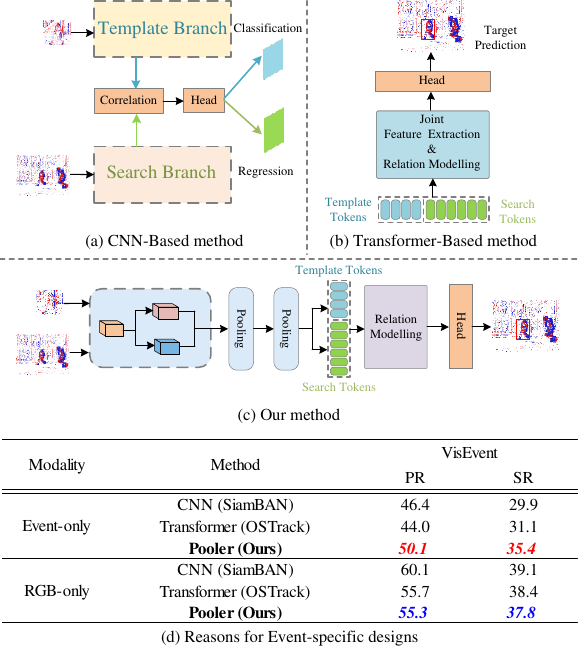}
\caption{Comparisons of the proposed Pooler with CNN and Transformer methods. The CNN-based method uses parallel branches to extract the template and search features. The Transformer-based method utilises a unified attention block to perform relation modelling of the template and search tokens. Our pipeline takes into account the sparsity property of event images. ``Event-only" and ``RGB-only" signify that the input contains only one specific modality.}
\label{fig:comparison}
\end{figure}

Visual object tracking aims to detect and locate  the target, specified in the initial frame of a video, in the subsequent video frames. 
It supports a wide range of applications, such as automatic driving, surveillance, UAV navigation~\cite{wen2023enhanced}. The existing technology based on RGB images is impressive in good imaging conditions. However, RGB images can be degraded by, for instance, limited illumination, or over-exposure. 
To address these issues, the current tracking community has been investigating the merits of event data~\cite{gallego2020event, zheng2023deep, tang2022periodic}, that can heighten the awareness of the changing light intensity, in the context of robust visual tracking. 
Although event data offers great advantages in perceiving motion, it is unable to capture the visual appearance of the target object, such as colour and texture information. 
Therefore, visual object tracking, combining RGB and event data has been gaining increasing attention.

Unfortunately, the inherent disparities between the RGB and event data modalitites pose notable challenges in devising effective strategies for the complementary use of the two modalities.
The two primary challenges are: 
(1) The effective extraction of robust event features. The discrepancies between RGB and event sensors pull their distributions to different domains. However, existing trackers process the sparse event data using the tools borrowed from the RGB data analysis, yielding a sub-optimal solution. (2) How to harness the advantages of both modalities. Notably, an advanced tracker, ViPT~\cite{ViPT} introduces a strategy in which a pre-trained RGB model is fine-tuned~\cite{bahng2022exploring} to learn event-related prompts~\cite{jia2022visual, yang2022prompting}, resulting in commendable performance. However, it still employs a simple addition operator to aggregate the features from both modalities, failing to achieve effective fusion of the information from the two modalities. 

To address the aforementioned challenges, a novel end-to-end RGB-E tracking network is proposed. By taking into account the sparsity of event data, we can extract  high-quality event features through a dedicated design of an event feature extraction backbone. 
The complementary information conveyed by the two modalities can effectively be fused by each modality focusing on the data it is competent to extract. %thereby augmenting their respective features. 
Specifically, (1) we introduce a novel Pooling-based Event backbone that prioritises the position and shape of the moving object. 
The sparsity characteristic of event images causes some regions to have high pixel values, compared to others, while other regions assume zero pixel values. 
The pooling operation is conceived to concentrate on these sparse pixel values as a way to retain the crucial information. 
As shown in Figure~\ref{fig:comparison}(d), when the input is just the event modality, our Pooler outperforms the conventional CNN and Transformer solututions, which are typically employed in the current practices. 
In contrast, on the data captured by the  RGB modality, the performance of our Pooler falls short, compared to the two established solutions. 
These experiments conclusively support the basic premise of our approach to handling event data. 
Our Event-specific Pooler excels in effectively extracting and capturing high-quality event features, setting it apart from traditional RGB backbones.
(2) To enable effective information fusion of both modalities, we introduce a Mutually-Guided Fusion module (MGF). 
Our MGF allows each modality to inject the relevant information from the other source using a cross-attention mechanism, thereby improving their feature representation. The results of extensive experiments conducted on various RGB-E datasets
 validate our proposed TENet by significantly outperforming existing state-of-the-art RGB-E tracking methods.

In conclusion, our contributions can be summarised as follows:
% \begin{figure}[t]
% \begin{center}
% \includegraphics[width=0.32\linewidth]{./img/101.png}
% \includegraphics[width=0.32\linewidth]{./img/102.png}
% \includegraphics[width=0.32\linewidth]{./img/103.png}\\
% \includegraphics[width=0.32\linewidth]{./img/201.png}
% \includegraphics[width=0.32\linewidth]{./img/202.png}
% \includegraphics[width=0.32\linewidth]{./img/203.png}\\
% \includegraphics[width=0.32\linewidth]{./img/301.png}
% \includegraphics[width=0.32\linewidth]{./img/302.png}
% \includegraphics[width=0.32\linewidth]{./img/303.png}\\
% \includegraphics[trim={0mm 126mm 170mm 5mm},clip,width=0.7\linewidth]{./img/frtt_legend.pdf}
% \end{center}
% \caption{Illustration of the qualitative tracking results on challenging hazy sequences (the listed videos are \textit{Haze-Bolt}, \textit{Haze-Ironman}, and \textit{Haze-Bike} from the Haze-OTB2015 dataset). The colour bounding boxes are the corresponding predictions of D3S, SiamRPN++, DiMP, ATOM, and the proposed FRTT tracker.}
% \label{qualitative}
% \end{figure}
\begin{itemize}
\item A novel, lightweight Pooling-based event feature extraction backbone that incorporates a multi-scale pooling operation to extract informative event features. The dedicated Pooler is instrumental in preserving motion clues and target contour, while ignoring the target  appearance.

\item A cross-attention based Mutually-Guided Fusion module, which enables both modalities to concentrate on features that are prominent in their respective sensing domains, and fuses them effectively.

\item An extensive experimental validation demonstrates that our proposed TENet surpasses the performance of state-of-the-art trackers on the VisEvent and COESOT datasets, confirming the merit of the proposed fusion of the appearance and motion information.
\end{itemize}

\section{Related Works}
\label{related_work}

\subsection{Visual Tracking Frameworks}

Recently, visual object tracking~\cite{SiamFC, xu2019learning} has made tremendous progress thanks to the relentless advances in deep learning~\cite{schmidhuber2015deep}. 
The online discriminative trackers~\cite{paul2022robust, xu2021adaptive} learn a classifier to locate the object from the background.
A landmark in tracking has been created by the adoption of Siamese-based trackers~\cite{SiamRPN, SiamRPN++, xu2023toward}, incorporating a regional proposal network structured around the Siamese architecture and its design variations, aiming to enhance the features used for tracking. 
The Transformer-based trackers~\cite{ostrack, stark, xie2022correlation} employ the attention mechanism~\cite{Transformer} to capture long-term relational dependency between the template and the search region, promoting better accuracy and efficiency in long term visual object tracking.

While advanced models exhibit excellent tracking performance, a common characteristic among most models is their reliance on comparing templates with semantically similar search regions. 
This process of matching is inevitably affected by the instantaneous imaging characteristics of the input pairs. 
In specific scenarios, such as overexposure and high-speed motion, where image quality is compromised, it significantly degrades the model performance. 
Consequently, researchers are keen to integrate multi-modal~\cite{tang2023exploring, tang2023generativebased, zhu2023rgbd1k} inputs to mitigate  the limitations of a single imaging modality. 
The incorporation of the event modality in object tracking offers the means of providing complementary information that should enhance  the model's adaptability to complex real-world scenarios.

\subsection{RGB-E Tracking Solutions}
RGB-E tracking~\cite{zhang2023frame, chen2020end} becomes increasingly popular due to the superiority of event data in perceiving motions, providing high-precision dynamic timestamps. 
This multimodal tracking proves highly effective in dealing with fast-moving objects, with notable resilience to challenging illumination conditions. 
To benefit from the complementary sources of information, Zhang~\textit{et al}.~\cite{FE108} introduced a cross-domain feature integrator that adeptly fuses feature information from the two domains, utilising a cross-domain attention mechanism. 
They also developed a voxel-based event pre-processing approach and constructed an extensive event-based dataset. 
Following the fusion study, Wang~\textit{et al}.~\cite{Visevent} presented a cross-modal Transformer module designed to integrate Event data and RGB data.
To exploit the highly dynamic nature of event data, AFNet~\cite{AFNet} harnessed the elevated temporal resolution of event data to achieve high frame rate tracking. 
The model incorporates a cross-modal style alignment module, a cross-frame rate alignment module, and a cross-correlation fusion structure, facilitating a comprehensive RGB-E fusion.
Zhu~\textit{et al}.~\cite{HRCeutrack} employed the concept of Masked AutoEncoder (MAE~\cite{he2022masked}) to selectively mask RGB tokens and event tokens, enhancing interactions between cross-modal tokens. 
Additionally, they incorporated orthogonal high-rank regularisation to mitigate the network fluctuations induced by the masking process. 
To enable non-RGB to RGB transfer learning, ViPT~\cite{ViPT} presents a comprehensive framework for multi-modal tracking, incorporating spatial attention to model interactions between RGB tokens and non-RGB prompts. 
This methodology fine-tunes a pre-trained base model to improve performance. 
Their approach demonstrates the considerable potential of leveraging diverse visual cue learning in the realm of multi-modal tracking.

From the above overview it is apparent that  existing studies employ a homogeneous shared-weight backbone for the extraction of features by both the event and RGB modalities. 
This overlooks the divergent characteristics of the event and RGB data that are evident from the nature of their acquisition processes: event data, captured by an event camera, records asynchronous brightness changes within the scene at an irregular frame rate, while RGB data, obtained through an RGB camera, captures images at a consistent and fixed frame rate. 
We argue, therefore, that it is crucial to design separate, modality specific processing for the event modality to effectively exploit its unique properties.
As the process of fusion adopted by the above methods is handcrafted, it will also be necessary to revisit the methodology of fusing the two modalities to ensure that the complementary characteristics of these modalities are properly integrated. 

Typically, the event modality exhibits sparsity, with only a limited number of events being detected from time to time. To reflect these characteristics, we design a novel Pooling-Based architecture, devising an extraction backbone for event features that capture the motion information from adjacent regions. 
% This strategy aims to preserve the primary motion features of the event data. 
% Our Pooler is devised to address the first issue. 
On the multi-modal fusion side, we integrate a cross-attention mechanism, enabling each modality to selectively focus on features relevant to its specific domain. 
We shall demonstrate that this design facilitates a mutual reinforcement of domain specific features. 
As the attention mechanism within a Transformer framework excels at dynamically capturing dependencies across diverse modalities, 
 its investigation will be at the top of the list of options for amalgamating the multi-modal information.

\begin{figure*}[t]
\begin{center}
\includegraphics[width=1\linewidth]{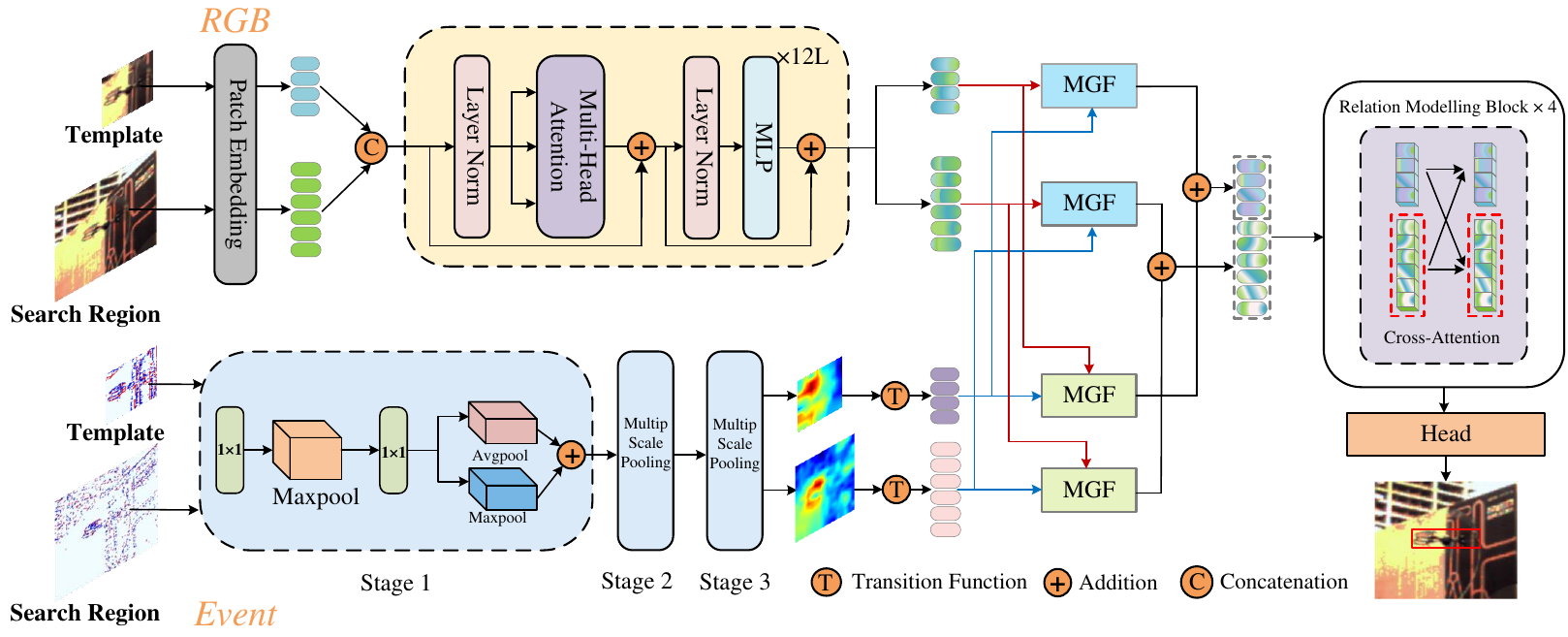}
\caption{{(a) Overview of our TENet, The 3 Stages of the Event branch collectively constitute the designed event backbone ``Pooler".} 
}
\label{fig:1}
\end{center}
\end{figure*}

\section{The Proposed Method}
\subsection{Network Overview}
The overall architecture of our TENet is shown in Figure~\ref{fig:1}. Our Network contains four components: Two-modality asymmetric backbone, Mutually-Guided Fusion module (MGF), the relation modelling module, and the head. Given the inputs, we first utilise the transformer backbone to extract RGB features and employ our Pooler to extract event features. Subsequently, we apply the MGF to promote mutual reinforcement of the features from both modalities. The features from the template and the search region of the two modalities are progressively fused. Next the fused template features and search region features are concatenated to facilitate the relation modelling. Finally, the acquired tokens are input to the tracking head for object localisation.

\subsection{Pooling-based event feature extraction backbone}

\begin{figure}[t]
\begin{center}
\includegraphics[width=0.65\linewidth]{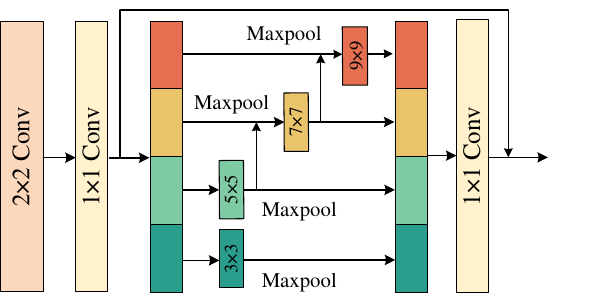}
\caption{{Details of Multi-Scale Pooling module.} 
}
\label{fig:2}
\end{center}
\end{figure}

Our Pooler is designed to extract high-quality event features for subsequent fusion with the features of both modalities. Our design takes into account the event data properties. It differs from the RGB modalities in several respects. Notable distinctions are its sparsity and its heightened sensitivity to motion. These properties motivate the use of a Pooling operation for extracting the motion information from event data, serving as a valuable complement to the target appearance provided by the RGB data. 
Given that some pixels in sparse event images are of high value, the Pooling operation is able to focus on such regions, while averaging out the noise and insignificant changes.
Moreover, by adopting a Multi-Scale pooling, our feature aggregation captures different feature scales.
The multi-level approach enhances the robustness of representations for sparse features by capturing fine-grained details at smaller scales and encapsulating abstract information at larger scales, thereby enabling effective handling of various feature distributions. 
Traditional convolution operations are dedicated to extracting detailed semantic features of the object, whereas event-based modalities concentrate on capturing the dynamic aspects of object movement. Thereby,our Pooler is more committed to a clear depiction of motion.
% \textcolor{red}{How, why?}

Our Pooler is divided into 3 stages. 
Event data often includes numerous events, but not all are task-relevant. In fact, there is a significant number of noise events. 
The role of the first stage is first to apply MaxPooling to the event data in order to identify crucial events and mitigate noise events so as  to capture salient feature representations. Subsequently, we employ AvgPooling to capture low-level information within the event data. %and MaxPooling is utilised to capture high-level information from the same data. 
In the final step, the gathered low-level and high-level event data are merged to create a more comprehensive event representation. The event features $F_{E_1}$ obtained in Stage 1 can be expressed as follows:

\begin{equation}
\left\{
\begin{aligned}
    &F_{E_f}={\emph M}({{\varphi}_{1\times1}}(E)), \\
    &F_{E_i}={\emph M}({{\varphi}_{1\times1}}(F_{E_f})),\\
    &F_{E_j}={\emph A}({{\varphi}_{1\times1}}(F_{E_f})),\\
    &F_{E_1}=F_{E_i}+F_{E_j},
\end{aligned}
\right.
\end{equation}

% \begin{equation}
% F_{E_f}={\emph M}({{\varphi}_{1\times1}}(E)),
% \end{equation}
% \begin{equation}
% F_{E_i}={\emph M}({{\varphi}_{1\times1}}(F_{E_f})),
% \end{equation}
% \begin{equation}
% F_{E_j}={\emph A}({{\varphi}_{1\times1}}(F_{E_f})),
% \end{equation}
% \begin{equation}
% F_{E_1}=F_{E_i}+F_{E_j},
% \end{equation}
where $E$ represents the original event images. $F_{E_f}$, $F_{E_i}$, and $F_{E_j}$ correspond to the features obtained during Stage 1. ${\varphi}_{1\times1}$  represents a ${1\times1}$ convolutional layer. ${\emph M}$ represents MaxPooling operation and ${\emph A}$ represents AvgPooling operation.

\begin{figure}
\centering
\includegraphics[width=1\linewidth]{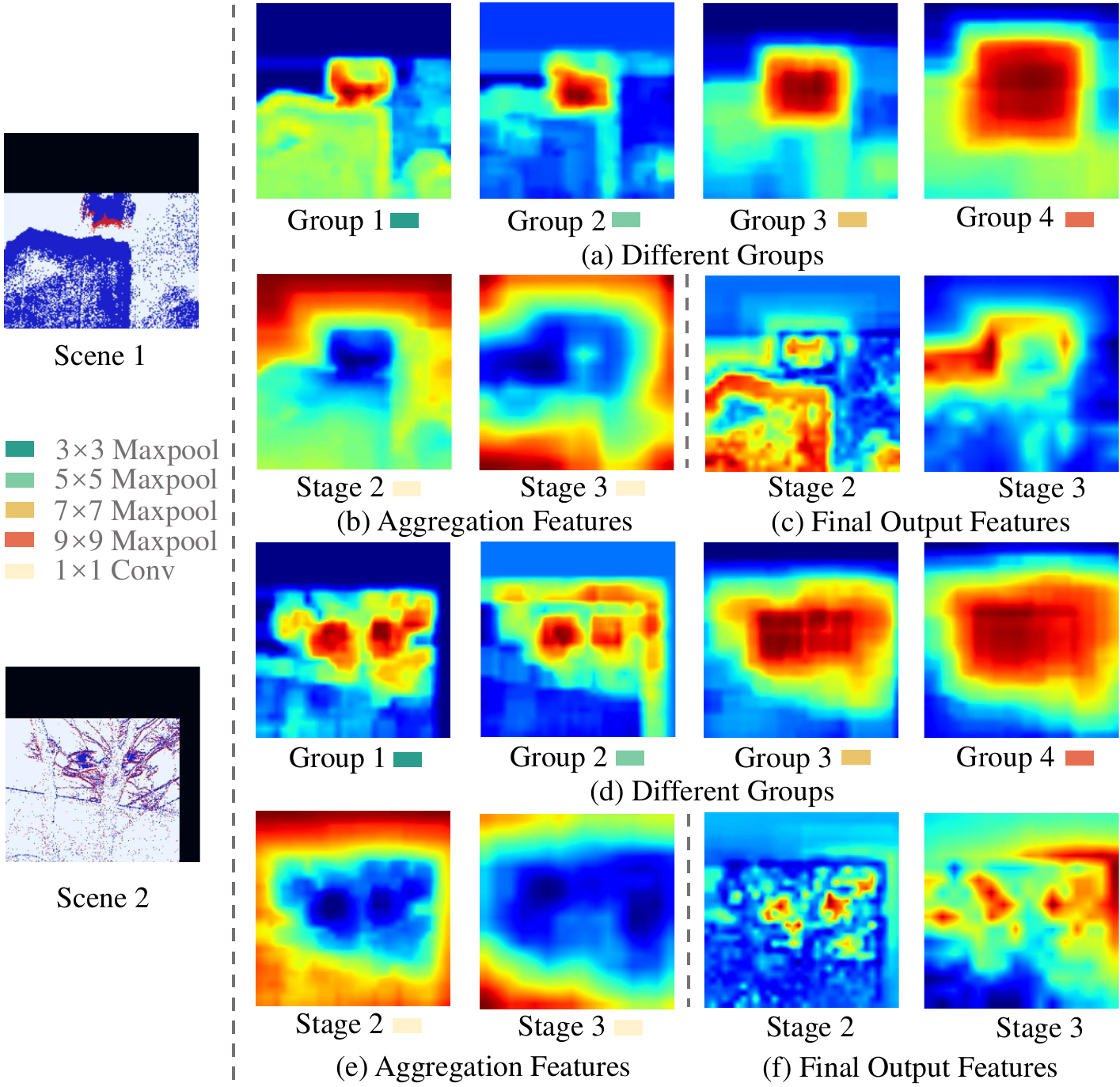}
\caption{Visualisation of the output features of different groups in the Multi-Scale Pooling during Stage 2. Group $n$ indicates that the $n$-th in the 4 divided groups. $n={1,2,3,4}$. (b) and (e) Visualisation of the aggregation features from Stage 2 and Stage 3. (c) and (f) Visualisation of the final output features from Stage 2 and Stage 3.}
\label{fig:feature}
\end{figure}

The event camera captures events with a microsecond-level latency, providing real-time responses to rapid changes in high-speed scenarios and tracking the changes in object position and location. 
To effectively handle objects appearing at different positions and moving with distinct speeds, we introduce Multi-Scale Pooling (MSP) operations in Stage 2 and Stage 3. 
MSP adjusts the scale of the analysis based on the variations in object velocity to capture the changes in the object more accurately. 
This module utilises pooling operations with different kernel sizes, allowing the network to achieve a range of feature representations that comprehensively capture motion, from coarse to fine-grained detail.

More specifically, in Stage 2, as depicted in Figure~\ref{fig:2}, we partition the features produced in Stage 1 into four groups based on their channel dimensions. 
The first group is initialised with a ${3\times3}$ kernel, while the kernel size for each subsequent group is increased by 2. 
This approach permits each group to possess a unique receptive field, thereby effectively capturing changes in both the object and the background. 
Subsequently, as each group contains information of varying granularity, we apply a ${1\times1}$ convolution to all groups to aggregate the extracted information across groups, promoting global information cross-fertilisation. 
The feature $F_{E_2}$ obtained in Stage 2 can be expressed as follows:

\begin{equation}
F_{E_i}={{\varphi}_{1\times1}}({{\varphi}_{2\times2}}(F_{E_1})),
\end{equation}
\begin{equation}
\begin{aligned}
& F_{E_j}=\emph Concat(M_{{k_1}\times{k_1}}(\textit x_1), M_{{k_2}\times{{k_2}}}(\textit x_2), \\
& M_{{k_3}\times{{k_3}}}(\textit x_3+G_2 ), ... , M_{{k_n}\times{k_n}}(\textit x_n+G_{n-1} )),
\end{aligned}
\end{equation}
\begin{equation}
F_{E_2}={\varphi}_{1\times1}(F_{E_j})+F_{E_i},
\end{equation}
where $F_{E_i}$ and $F_{E_j}$ denote the features obtained in Stage 2. $x=\{x_1,x_2,...,x_n\}$ refers to the feature of each group along the spatial dimension of $F_{E_j}$, which is partitioned into $n$ groups. $k_i=\{3,5,...,K\}$ denotes the kernel size of each group, which is gradually increased by 2. $G_n$ signifies the result of the $n$-th group processing, following the MaxPooling operation.

For intuitive visualisation, we select two representative scenes to validate the effectiveness of our Pooler. 
As shown in Figure~\ref{fig:feature}(a) and Figure~\ref{fig:feature}(b), four different scales of pooling operations are utilised to focus on objects in event images in both, Scene 1 and Scene 2. 
Groups 1 and 2 in both scenes are capable of concentrating on fine-grained objects and the surrounding environment. 
Groups 3 and 4 experience a certain degree of feature loss as a result of the use of a large convolutional kernel in the pooling operation, but this helps to capture coarse features, that are spatially imprecise. 
The features extracted at each scale are then aggregated to acquire a comprehensive representation. % the features after the pooling operation at four different scales. 
From Figure~\ref{fig:feature}(b) and Figure~\ref{fig:feature}(e), the variations in objects at various scales, along with background elements, are aggregated to make it possible to segregate the object from the background. From both Stage 2 and Stage 3 in Figure~\ref{fig:feature}(c) and Figure~\ref{fig:feature}(f), our MSP successfully differentiates the object from the background, thereby effectively extracting event features.

\subsection{Mutually-Guided Fusion module}

\begin{figure*}[t]
\begin{center}
\includegraphics[width=1\linewidth]{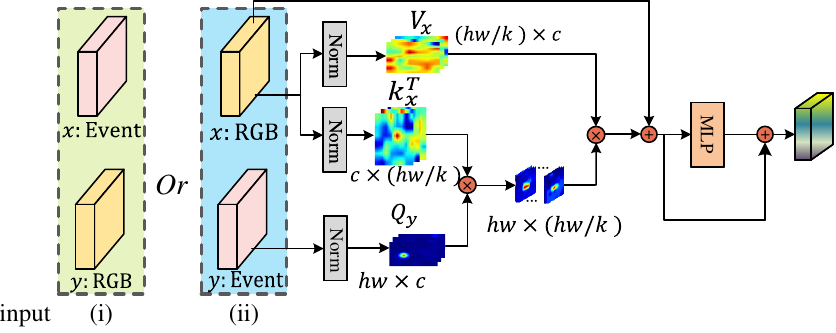}
\caption{{Details of the proposed Mutually-Guided Fusion (MGF) module. The input to MGF is (i) or (ii). (i) denotes the augmentation of the Event features through the inclusion of RGB-related features. (ii) denotes the augmentation of RGB features with the inclusion of the Event-related features.} 
}
\label{fig:3}
\end{center}
\end{figure*}

In general, the fundamental task of RGB-E tracking lies in the synergistic fusion of event motion cues and RGB appearance information. 
% In existing approaches, ViPT~\cite{ViPT} involves a linear mapping of the two-modality data to low-quality tokens \textcolor{red}{This statement is objectionable. Why should RGB token be of low quality? What is of low quality and why?} and subsequently employing a basic addition fusion. 
In existing approaches, ViPT~\cite{ViPT} involves a linear mapping of the two-modality data to tokens and subsequently employing a basic addition fusion. 
However, this straightforward fusion method falls short in unlocking the complete potential of event and RGB data, overlooking their unique advantages. 
Other methods design the model structure manually, taking the specific Event data characteristics into account, but this is likely to be less effective than a solution learnt by a purposeful fusion architecture. 

To enhance the utilisation of event data, we introduce a mutually guided cross-modal fusion module into our model, namely MGF. 
This module leverages cross-attention mechanisms, enabling the salient features of one modality to integrate with those of the other, thereby strengthening each modality and fostering a more cohesive inter-modal collaboration. 
To minimise the spatial complexity associated with the aforementioned attention operations, we downsize the width and height of keys and values to the $k$-th percentile of their original sizes. 
Specifically, the attention map obtained in the original Transformer \cite{Transformer} according to \eqref{equ1} is $hw \times hw$. 
Following the downsampling operation, the dimensions of the attention map are altered to $hw \times (hw/k)$. 
In input (ii) in Figure~\ref{fig:3}, we show the process of enhancing the RGB modality with the event modality as an example. 
The input features from the two branches are initially subjected to layer normalisation, followed by applying the cross-attention mechanism to process the tokens from both modalities. 
In this process, the vectorised features of the event modality are used as queries $Q_{Event}$ and the vectorised features of RGB modality are used as keys $K_{RGB}$ and values $V_{RGB}$, respectively. This process can be represented by the following equation:
\begin{equation}
\label{equ1}
{\emph Attention}(Q_{Event},K_{RGB},V_{RGB})= softmax(\frac{Q_{Event}{(K_{RGB})^T}}{d_k})V_{RGB},
% & Attention(Q_{Event},K_{RGB},V_{RGB})= \\
% & softmax(\frac{Q_{Event}{K_{RGB}}^T}{d_k})V_{RGB}
% & {\emph Attention}(Q_{Event},K_{RGB},V_{RGB})= \\
% & softmax(\frac{Q_{Event}{K_{RGB}}^T}{d_k})V_{RGB},
% \begin{aligned}
%     & {\emph Attention}(Q_{Event},K_{RGB},V_{RGB})= \\
%     & softmax(\frac{Q_{Event}{K_{RGB}}^T}{d_k})V_{RGB}    
% \end{aligned}
\end{equation}
where $d_k$ is the dimension of keys($K_{RGB}$). This formula describes how to use query tokens of the event modality to pay attention to the key tokens of the RGB modality, in order to focus on the associations of these two modalities, and then use RGB value tokens to further enhance the similarities. 
Finally, the original RGB tokens are added to the output of the attention operation. 
The combined result is then passed on to the MLP for further processing. 
Enhancing the event modality with the RGB modality follows a similar process, with the key difference being that it involves swapping the roles, namely utilising the vectorised features of the RGB modality as queries ($Q_{RGB}$) and utilising the vectorised features of the event modality as keys ($K_{Event}$) and values ($V_{Event}$).

\subsection{Relationship modelling for fusing template and search}
Following the fusion of features from the RGB and event modalities, the subsequent step involves independently summing up the template and search region tokens for each of these modalities. 
This operation serves to preserve the independence of the two modalities while efficiently integrating their information. 
Prior to modelling the feature relationships, we concatenate the template tokens and search region tokens along the spatial dimension to create a unified representation. 
Subsequently, we use four layers of the standard ViT~\cite{vision} architecture to conduct relationship modelling on the combined tokens. 
At each layer, we utilise a Multi-Head Self-Attention (MHSA) module to calculate self-attention among the combined tokens. 
This involves computing the cross-attention between the template tokens and the search region tokens, after combining the two modalities to capture their interrelationships. 
Finally, a split operation is applied to separate the search region tokens for the purpose of object localisation. The overall process of relationship modelling can be depicted as follows:
\begin{equation}
\left\{
\begin{aligned}
    & F_T=fu_{E\_T}+fu_{R\_T},\\
    & F_S=fu_{E\_S}+fu_{R\_S},\\
    & {f_l}^0=Concat(F_T,F_S), \\
    & {{f_l}^{i'}}={{f_l}^{i}}+MSA(LN({{f_l}^{i}})),\\
    & {{f_l}^{i+1}}={{f_l}^{i'}}+MLP(LN({f_l}^{i'})), \\
    & {{final_S}^I}=DeConcat(f^I),
\end{aligned}
\right.
\end{equation}

% \begin{equation}
% F_T=fu_{E\_T}+fu_{R\_T}, F_S=fu_{E\_S}+fu_{R\_S},
% \end{equation}
% \begin{equation}
% {f_l}^0=Concat(F_T,F_S), 
% \end{equation}
% \begin{equation}
% {{f_l}^{i'}}={{f_l}^{i}}+MSA(LN({{f_l}^{i}})),
% \end{equation}
% \begin{equation}
% {{f_l}^{i+1}}={{f_l}^{i'}}+MLP(LN({f_l}^{i'})), 
% \end{equation}
% \begin{equation}
% {{final_S}^I}=DeConcat(f^I),
% \end{equation}
where $l$ stands for the $l$-th layer and $I$ for the last layer. 

\section{Evaluation}
\subsection{Implementation details.} 
In our approach, we employ the feature extraction backbone for the RGB modality~\cite{ostrack}, which has been pre-trained on tracking datasets. 
Our implementation is carried out using Python 3.7 and PyTorch 1.9.0. 
The network is trained over 60 epochs, utilising the AdamW optimiser with default settings. 
During training, we use a batch size of 32, and the initial learning rate is set to 0.0001. 
The network training is conducted on a single RTX 3090 GPU.

\subsection{Comparison with the state-of-the-art Trackers}
We evaluate the performance of our network on two RGB-E benchmarks, namely VisEvent and COESOT. 
For evaluation, we use precision rate (PR) and success rate (SR) as the measurement metrics. 
It is worth noting that both datasets were recorded using a DAVIS 346 camera with a resolution of 346 × 230 pixels. 
Importantly, we exclusively use event images derived from the original event data and do not utilise event streams.

\begin{figure}
  \begin{minipage}{0.5\linewidth}
    \centering
    \includegraphics[width=1\linewidth]{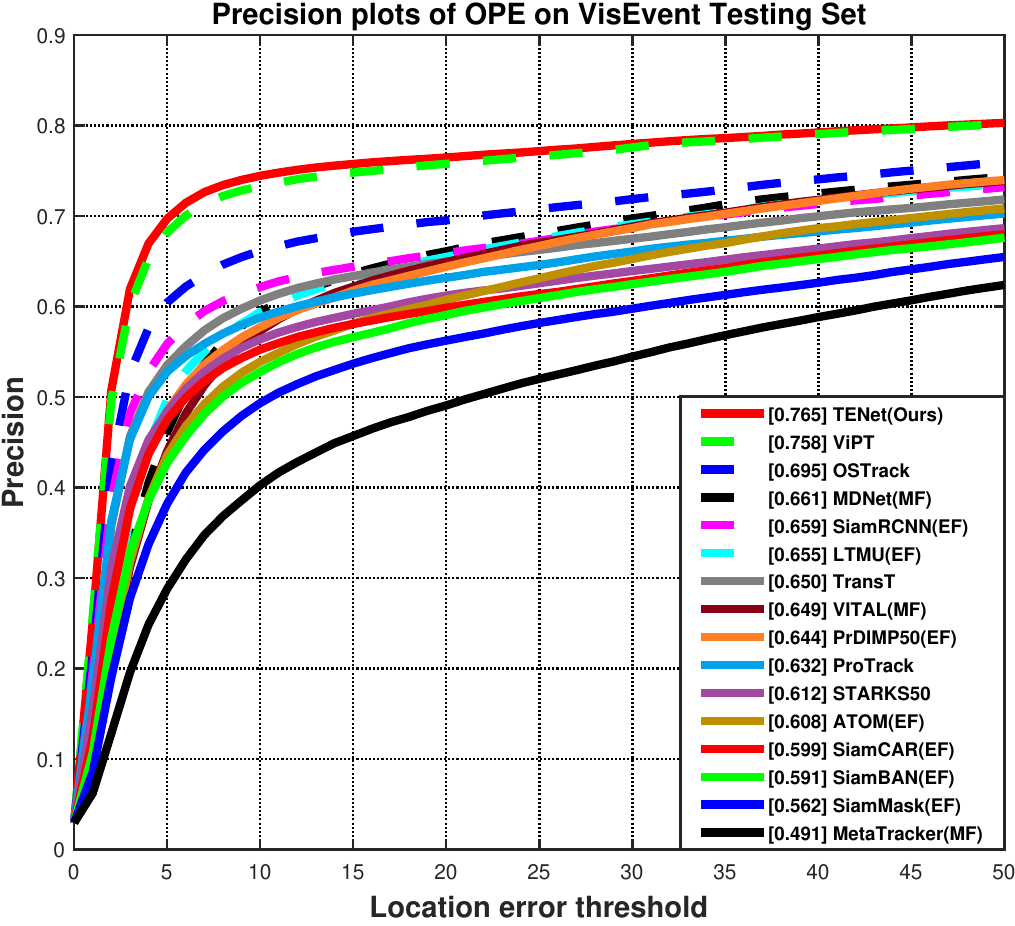}
  \end{minipage}\hfill
  \begin{minipage}{0.5\linewidth}
    \centering
    \includegraphics[width=1\linewidth]{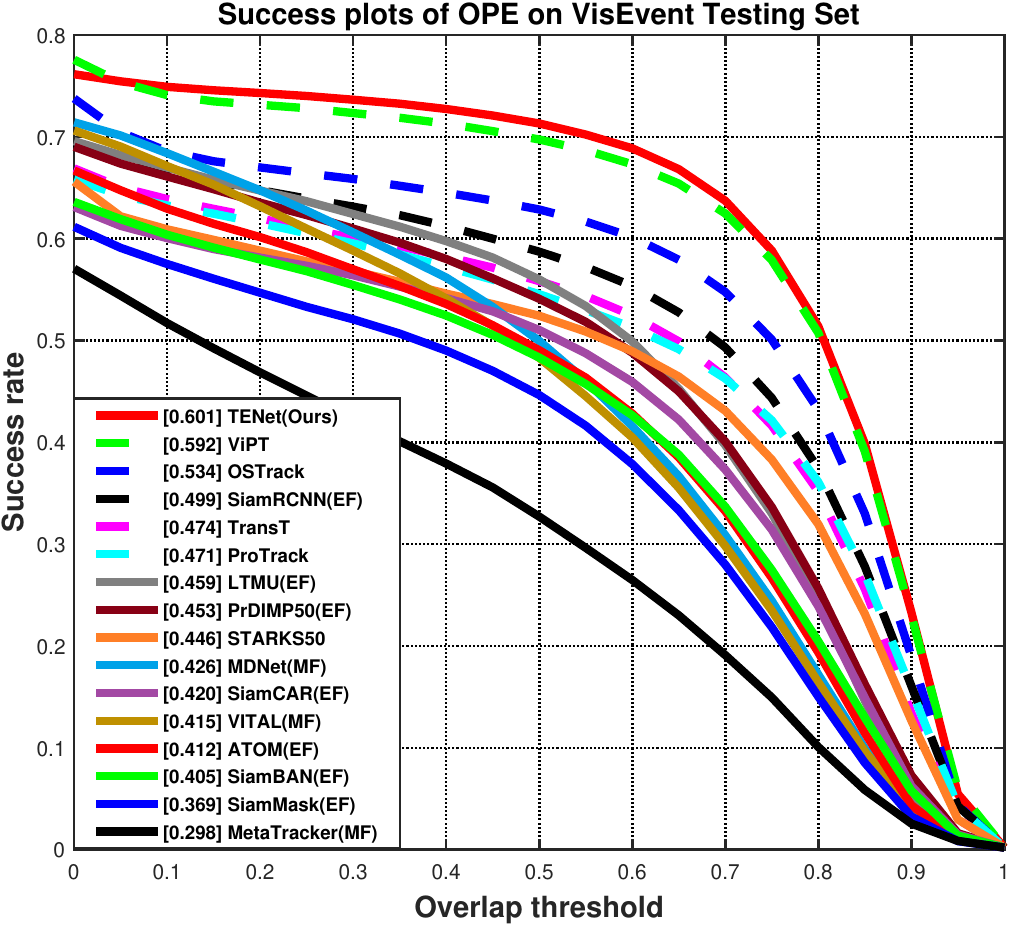}
  \end{minipage}
  \caption{Precision and success plots of the VisEvent benchmark.}
  \label{fig:4}
\end{figure}

\textbf{Results on VisEvent.} VisEvent is presently among the extensively utilised datasets, comprising 820 pairs of videos captured in environments characterised by low illumination, high dynamics, and background clutter. The dataset is partitioned into 500 training subsets and 320 test subsets. As depicted in Figure~\ref{fig:4}, in comparison to the state-of-the-art tracker ViPT~\cite{ViPT}, our method enhances the precision rate (PR) and success rate (SR) by 0.7\% and 0.9\% to reach 76.5\% and 60.1\%, respectively. Notably, our approach achieves an inference speed that is approximately 40\% faster than ViPT. Both the experimental results and the improved inference speed underscore the effectiveness and efficiency of our method.

\textbf{Results on COESOT.} COESOT stands as a generic single object tracking dataset designed for colour event cameras, containing 1354 colour event videos and 478,721 RGB frames. 
The dataset is categorised into 827 training subsets and 527 testing subsets. Table~\ref{tab:coe} reports the results of a comprehensive evaluation of our method on the COESOT dataset, showing a commendable performance in both accuracy and success rate. 
Specifically, our proposed method attains a precision rate (PR) of 76.8\% and a success rate (SR) of 68.4\%, surpassing the state-of-the-art method, HRCeutrack, by 4.9\% and 5.2\%, respectively. 
This represents a substantial performance enhancement in the event tracking task. 
% To validate the robustness, we subject it to testing across 17 common challenges.
% Our method outperforms all other approaches in these common challenges, affirming its exceptional adaptability to diverse tracking challenges. 
% The results of 17 challenges can be referred to the Fig.~\ref{fig:challenges}.
% Our TENet achieves the best performance on all of these challenges, like background clutter, background object motion, camera motion, full occlusion, viewpoint change and deformation, etc. 
% These results collectively demonstrate that our method not only exhibits outstanding generalisation ability but also shows efficiency across various challenges.

% \begin{figure*}[t]
% \begin{center}
% \includegraphics[width=1 \linewidth]{17_challenges.pdf}
% % \includegraphics[width=1 \linewidth]{111.pdf}
% %,height=4.5cm
% \caption{Precision and success plots of trackers on 17 challenges.} 
% \label{fig:challenges}
% \end{center}
% \end{figure*}

\begin{table}[!t]
    \caption{{An overall comparison on the COESOT benchmark. We adopt four commonly used metrics for the comparison, i.e., PR, SR, normalised precision rate (NPR), and breakOut capability score (BOC). ``EI" stands for event images. ``EVox" represents the voxel form of the raw event data.}}\vspace{2mm}
    \centering
    \scriptsize
		\begin{tabular}{cc|c|cccc}
            \toprule
            & Models & Modality & SR & PR & NPR & BOC \\\midrule\midrule
            
            & Mixformer1k~\cite{mixformer} & RGB+EI     &56.0   & 62.8 & 61.7 & 17.2 \\
            & PrDiMP50~\cite{prdimp}    & RGB+EI     & 57.9 & 65.0   & 64.0   & 17.5 \\
            & KYS~\cite{kys}         & RGB+EI     & 58.6 & 66.7 & 65.7 & 17.9 \\
            & DiMP50~\cite{DiMP}      & RGB+EI     & 58.9 & 67.1 & 65.9 & 18.1 \\
            & AiATrack~\cite{aiatrack}    & RGB+EI     &59.0   & 67.4 & 65.6 & 19.0 \\
            & KeepTrack~\cite{keeptrack}   & RGB+EI     & 59.6 & 66.1 & 65.1 & 18.1 \\
            & TOMP50~\cite{tomp}      & RGB+EI     & 59.8 & 66.7 & 65.7 & 18.3  \\
            & TrDiMP~\cite{trdimp}      & RGB+EI     & 60.1 & 66.9 & 65.8 & 18.4 \\
            & SuperDiMP~\cite{superdimp}   & RGB+EI     & 60.2 &67.0   &66.0   & 18.5 \\
            & TransT50~\cite{transt}    & RGB+EI     & 60.5 & 67.9 & 66.6 & 18.5  \\
            & SiamR-CNN~\cite{siamrcnn}   & RGB+EI     & 60.9 & 67.5 & 66.3 & 19.1 \\
            & CEUTrack~\cite{coesot}    & RGB+EVox &62.0   & 70.5 &69.0   & 20.8 \\
            & HRCEUTrack~\cite{HRCeutrack}  & RGB+EVox & 63.2 & 71.9 & 70.2 & 21.6 \\
            & \textbf{TENet (Ours)}        & RGB+EI     
            &\textbf{68.4} & 
            \textbf{76.8} & 
            \textbf{75.3} & 
            \textbf{24.2} \\
            \bottomrule
        \end{tabular}
    \label{tab:coe}
\end{table}

\subsection{Discussion of the experiments evaluating the  ``Pooler"}
\begin{figure*}[t]
\begin{center}
\includegraphics[width=0.9 \linewidth]{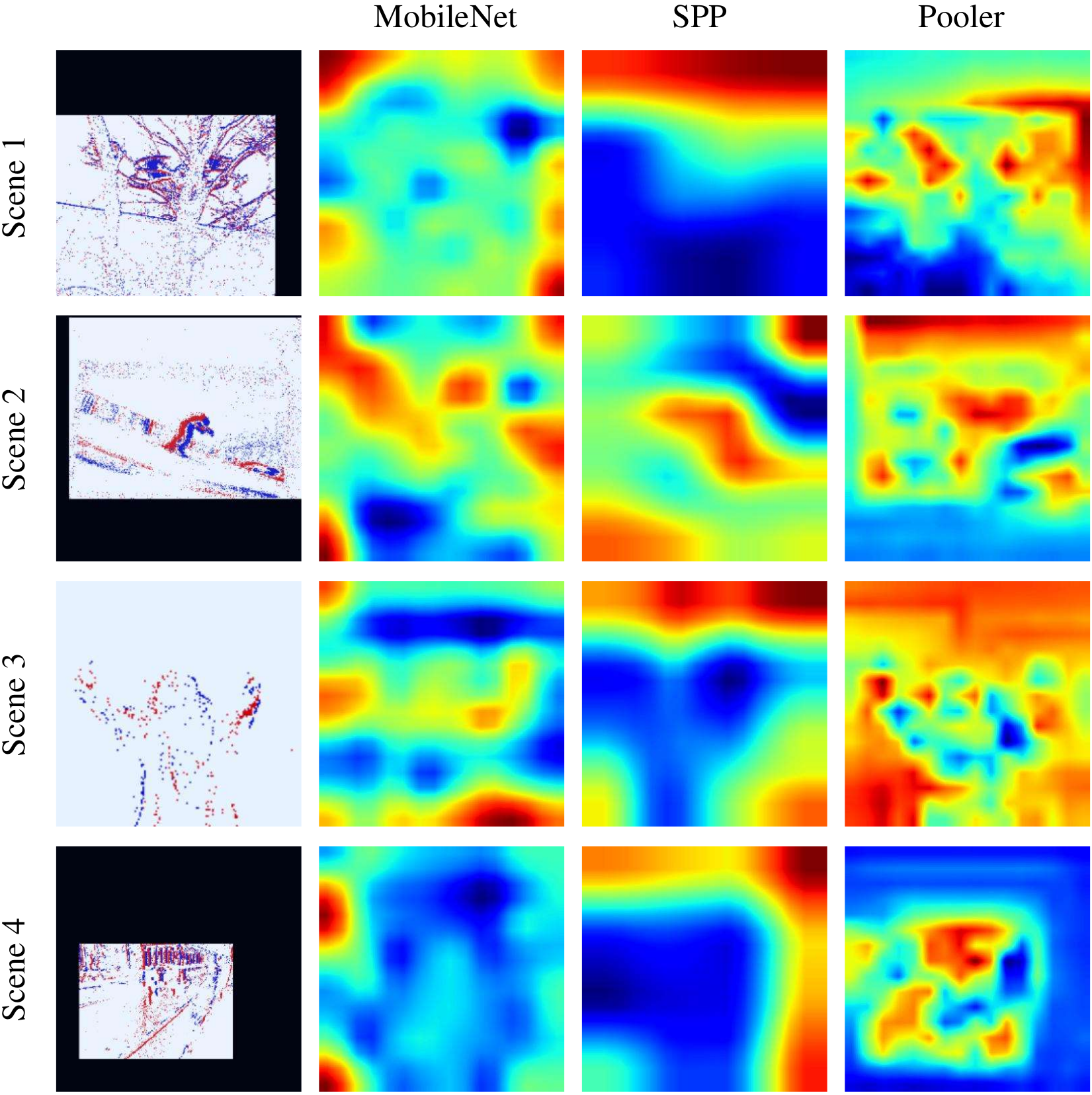}
\caption{Visualization of the feature maps of different methods in Table~\ref{tab:add}. SPP denotes Spatial Pyramid Pooling.} 
\label{fig:additional}
\end{center}
\end{figure*}

Based on the sparsity and discreteness of the data conveyed by the event frames, we propose a lightweight multi-scale pooling mechanism, termed Pooler, which  captures object motion information  across different scales. %thereby enhancing the feature extraction capability of event frames. 
In order to illustrate the advantages of Pooler over existing lightweight backbones and traditional multi-scale pooling methods, we compare its features with the output features of MobileNet~\cite{mobilenetv3} and SPP~\cite{he2015spatial} (Spatial Pyramid Pooling).
The configuration of the MobileNet and SPP competitor is illustrated in Figure~\ref{fig:1}. Basically, Pooler is replaced by MobileNetv3, and Stage2 and Stage3 in Pooler are substituted by SPP.
As shown in Table~\ref{tab:add}, overall, TENet delivers superior performance, compared to MobileNetv3 and SPP. In VisEvent, our approach outperforms MobileNetv3 by 1.5\% and 1.4\%, and surpass SPP by 2.4\% and 2.0\%. Similarly, on COESOT, our method is better than  MobileNetv3 by 1.1\% and 0.9\%, and exceeds SPP by 1.5\% and 1.3\%.
To gain intuitive understanding of the strength of the Pooler performance, we visualize the feature maps from sample scenes in Figure~\ref{fig:additional}, outputted by MobileNetV3, SPP and Pooler.
From the analysis of Scene 1, Scene 2, and Scene 4, it is evident that our Pooler is able to distinguish the foreground from the background effectively.
In Scene 1, even when the object is partially obscured, our Pooler highlights the unobscured parts of the object successfully, showcasing its robust tracking ability in complex occlusion scenarios.
In both Scene 2 and Scene 4, our Pooler again accurately delineates the foreground from the background. %while highlighting the object.
In Scene 3, the Pooler  highlights the object outline, facilitating a better targeted object feature extraction. This enhances the machine perception of the object shape and spatial structure.
In contrast, it is apparent that the lightweight MobileNet struggles to distinguish  the foreground from the background clearly, when handling background interference. It is challenging to discern the object explicitly from its feature maps. 
%In contrast, our lightweight Pooler demonstrates improved adaptability to feature variations in different scenes. This enhances the capability to distinguish between foreground and background more clearly.
Similarly, SPP tends to encounter difficulties in capturing the foreground, whereas our multi-scale Pooler, thanks to its receptive fields originating  from tapping the information conveyed by different channels, exhibits stronger performance.

\begin{table}[!t]
\centering
\caption{{Performance comparison with different event backbones. Method 3 is our TENet.}}\vspace{2mm}
\scriptsize
% \resizebox{0.88\columnwidth}{!}{
\begin{tabular}{c|ccc|c|c|cc|cc}
\hline
\multirow{2}{*}{Methods} & \multicolumn{3}{c|}{Event Backbone}                                                           & \multirow{2}{*}{FLOPS} & \multirow{2}{*}{FPS} & \multicolumn{2}{c|}{VisEvent}    & \multicolumn{2}{c}{COESOT}       \\ \cline{2-4} \cline{7-10} 
                         & \multicolumn{1}{c|}{MobileNet} & \multicolumn{1}{c|}{SPP} & \multicolumn{1}{l|}{Pooler} &                        &                      & \multicolumn{1}{c|}{PR}   & SR   & \multicolumn{1}{c|}{PR}   & SR   \\ \hline
1                        & \multicolumn{1}{c|}{\checkmark}         & \multicolumn{1}{c|}{}    &                             & 34.615 G                & 23.02 fps            & \multicolumn{1}{c|}{75.0} & 58.7 & \multicolumn{1}{c|}{75.7} & 67.5 \\
2                        & \multicolumn{1}{c|}{}          & \multicolumn{1}{c|}{\checkmark}   &                             & 36.374 G                & 37.77 fps            & \multicolumn{1}{c|}{74.1} & 58.1 & \multicolumn{1}{c|}{75.3} & 67.1 \\
3                        & \multicolumn{1}{c|}{}          & \multicolumn{1}{c|}{}    & \checkmark                           & 35.157 G               & 44.13 fps            & \multicolumn{1}{c|}{76.5} & 60.1 & \multicolumn{1}{c|}{76.8} & 68.4 \\ \hline

% \begin{tabular}{c|c|c|cc|cc}
% \hline
% \multirow{2}{*}{Methods} & \multirow{2}{*}{FLOPS} & \multirow{2}{*}{FPS} & \multicolumn{2}{c|}{VisEvent}    & \multicolumn{2}{c}{COESOT}       \\ \cline{4-7} 
%                          &                        &                      & \multicolumn{1}{c|}{PR}   & SR   & \multicolumn{1}{c|}{PR}   & SR   \\ \hline
% TENet + MobileNet        & 34.615 G               & 23.02 fps            & \multicolumn{1}{c|}{75.0} & 58.7 & \multicolumn{1}{c|}{75.7} & 67.5 \\
% TENet+SPP                & 36.374 G               & 37.77 fps            & \multicolumn{1}{c|}{74.1} & 58.1 & \multicolumn{1}{c|}{75.3} & 67.1 \\
% TENet(ours)              & 35.157 G               & 44.13 fps            & \multicolumn{1}{c|}{76.5} & 60.1 & \multicolumn{1}{c|}{76.8} & 68.4 \\ \hline
\end{tabular}
\label{tab:add}
\end{table}

\subsection{Ablation Study}
\textbf{The influence of the synergistic effects of the Event backbone (Pooler) and Mutually-Guided Fusion (MGF).} 
To validate the effectiveness of Pooler and MGF, ablation experiments are conducted on both datasets. Table~\ref{tab:event} presents the results of the experiments. In the absence of Pooler, we substitute the feature extraction backbones of both modalities with OSTrack~\cite{ostrack}, and share the weights during training. 

In Method 1, the absence of both, our Pooler and MGF, results in a noteworthy decline in performance. Specifically, VisEvent shows a decrease of 1.9\% and 1.9\% in Precision (PR) and Recall (SR), while COESOT exhibits a reduction of 1.5\% and 1.2\% in PR and SR, respectively. Moreover, removing Pooler and MGF leads to a 43.48\% increase in the computational effort of the model and a 22.11\% decrease in inference speed. 

Method 2, which retains our Pooler but discards MGF, results in a performance boost over Method 1. The model exhibits a significantly reduced computational effort, accompanied by a notable increase in inference speed. Pooler is designed to be sensitive to the sparsity of event images and has, therefore, the capacity to recognise that certain pixels within an event hold crucial information. Consequently, the informativeness of its features is much better than that of the event features obtained by the RGB backbone.

Method 3 retains MGF but removes Pooler, producing results which are better than those of Method 1. The Mutually-Guided Fusion (MGF) module models the dynamic relationship between the modalities by means of  an adaptive strategy. This is achieved by invoking cross-attention to selectively highlight features within a modality that are pertinent to the tracking task, thus enhancing the representation computed by the modality.
In Method 4, the retention of both, the Pooler module and the MGF module, produces the best results. The intricate coupling of the high-quality event features extracted by the Pooler module with the RGB features, promoted by our MGF module, results in a synergistic fusion that enhances the overall quality of representation.

\begin{table}[!t]
\caption{{Ablation study of the Pooler and MGF modules. ``\ding{51}" or ``\ding{55}" signifies whether this module is retained or removed in the experimental setup. ``\ding{51} Pooler" indicates that our Pooler is removed and instead, the backbone, (OSTrack~\cite{ostrack}), employed for the RGB modality, is used  for extracting features for the event modality. Method 4 is our TENet.}}\vspace{2mm}
\centering
\resizebox{0.85\columnwidth}{!}{
\begin{tabular}{c|c|c|c|c|cc|cc}

\hline
\multirow{2}{*}{Methods} & \multirow{2}{*}{Pooler} & \multirow{2}{*}{MGF} & \multirow{2}{*}{Macs} & \multirow{2}{*}{FPS} & \multicolumn{2}{c|}{VisEvent}    & \multicolumn{2}{c}{COESOT}       \\ \cline{6-9} 
                         &                         &                      &                       &                      & \multicolumn{1}{c|}{PR}   & SR   & \multicolumn{1}{c|}{PR}   & SR   \\ \hline
1                        & \ding{55}                       & \ding{55}                     & 50.444 G              & 34.37 fps            & \multicolumn{1}{c|}{74.6} & 58.2 & \multicolumn{1}{c|}{75.3} & 67.2 \\
2                        & \ding{51}                       & \ding{55}                     & 31.565 G              & 51.03 fps            & \multicolumn{1}{c|}{75.0} & 58.8 & \multicolumn{1}{c|}{75.7} & 67.4 \\
3                        & \ding{55}                        & \ding{51}                    & 54.036 G              & 28.91 fps            & \multicolumn{1}{c|}{75.2} & 58.7 & \multicolumn{1}{c|}{75.8} & 67.6 \\
4                 & \ding{51}                       & \ding{51}                    & 35.157 G              & 44.13 fps            & \multicolumn{1}{c|}{76.5} & 60.1 & \multicolumn{1}{c|}{76.8} & 68.4 \\ \hline
\end{tabular}
}
\label{tab:event}
\end{table}

\textbf{The merit of MGF fusion.} We validate our Mutually-Guided Fusion module by selectively removing MGF(ii) or MGF(i) in TENet and presenting the results of retraining in Table~\ref{tab:MGF(ii)}. 
In method 3, the MGF(i) module is removed, and the MGF(ii) module is retained. There is a slight decrease in Precision (PR) and Recall (SR) across both datasets. The appearance information captured by the RGB features is combined with the motion information conveyed by the event features, contributing to the enhancement of object motion consistency.

In the case of Method 4, when we insert the MGF(i) module, with the MGF(ii) module absent, the PR and SR performance on VisEvent drops by 1.9\% and 1.5\%, respectively, while on COESOT, the PR and SR decrease by 2.3\% and 0.8\%, respectively.
This decline affirms that the fusion of event features with RGB features combines object appearance and event motion effectively, providing a substantial performance boost to the tracker. 
When both, the MGF (i) and MGF (ii) modules are removed, the tracker experiences a considerable drop in performance on both datasets. This decline underscores the merits of the mutual guidance and enhancement of the two modalities, emphasising their crucial role in amalgamating the RGB appearance information and event motion information to achieve better performance.

\begin{table}[!t]
\caption{{Ablation study of the Mutually-Guided Fusion module. ``MGF(ii)" denotes the use of event tokens as queries, while RGB tokens serve as both keys and values in the attention mechanism. ``MGF(i)" denotes the use of RGB tokens as queries, while event tokens serve as both keys and values.}}\vspace{2mm}
\centering
\scriptsize
% \resizebox{0.70\columnwidth}{!}{
\begin{tabular}{c|cc|cc|cc|cc}
\hline
\multirow{2}{*}{Methods} & \multicolumn{2}{c|}{MGF(ii)}    & \multicolumn{2}{c|}{MGF(i)}    & \multicolumn{2}{c|}{VisEvent}    & \multicolumn{2}{c}{COESOT}       \\ \cline{2-9} 
                        & \multicolumn{1}{c|}{w} & w/o & \multicolumn{1}{c|}{w} & w/o & \multicolumn{1}{c|}{PR} & SR & \multicolumn{1}{c|}{PR} & SR \\ 
                        \hline
                        1 & \multicolumn{1}{c|}{\checkmark} &  & \multicolumn{1}{c|}{\checkmark} &  & \multicolumn{1}{c|}{\textbf{76.5}} & {\textbf{60.1}} & \multicolumn{1}{c|}{\textbf{76.8}} & {\textbf{68.4}} \\
                        2 & \multicolumn{1}{c|}{}  & \checkmark & \multicolumn{1}{c|}{}  & \checkmark & \multicolumn{1}{c|}{75.0} & 58.8 & \multicolumn{1}{c|}{75.7} & 67.4 \\
                        3 & \multicolumn{1}{c|}{\checkmark} &  & \multicolumn{1}{c|}{}  & \checkmark  & \multicolumn{1}{c|}{76.1} & 59.5 & \multicolumn{1}{c|}{76.5} & 68.2 \\
                        4 & \multicolumn{1}{c|}{}  & \checkmark & \multicolumn{1}{c|}{\checkmark} & & \multicolumn{1}{c|}{74.6} & 58.6 & \multicolumn{1}{c|}{74.5} & 67.6 \\ \hline
\end{tabular}
% }
\label{tab:MGF(ii)}
\end{table}

\textbf{The effect  of downsampling  on the Mutually-Guided Fusion module.} 

\begin{table}[!t]
\caption{{An ablation study of the Relation Modelling block. ``RM block" denotes the Relation Modelling block.}}\vspace{2mm}
\centering
\scriptsize
% \resizebox{0.58\columnwidth}{!}{
\begin{tabular}{ccc|cc|cc}
\hline
\multicolumn{1}{c|}{}                          & \multicolumn{2}{c|}{RM block} & \multicolumn{2}{c|}{VisEvent}                                                                    & \multicolumn{2}{c}{COESOT}                                                                       \\ \cline{2-7} 
\multicolumn{1}{c|}{\multirow{-2}{*}{Methods}} & \multicolumn{1}{c|}{w/o}       & w      & \multicolumn{1}{c|}{PR}                                   & SR                                   & \multicolumn{1}{c|}{PR}                                   & SR                                   \\ \hline
\multicolumn{1}{c|}{1}                         & \multicolumn{1}{c|}{\checkmark}         &        & \multicolumn{1}{c|}{74.1}                                 & 58.1                                 & \multicolumn{1}{c|}{73.7}                                 & 65.1                                 \\
\multicolumn{1}{c|}{2}                         & \multicolumn{1}{c|}{}          & \checkmark      & \multicolumn{1}{c|}{76.5}                                 & 60.1                                 & \multicolumn{1}{c|}{76.8}                                 & 68.4                                 \\ \hline
\multicolumn{3}{l|}{Drops}                                                               & \multicolumn{1}{c|}{{\color[HTML]{FF0000} \textbf{-2.4}}} & {\color[HTML]{FF0000} \textbf{-2.0}} & \multicolumn{1}{c|}{{\color[HTML]{FF0000} \textbf{-3.1}}} & {\color[HTML]{FF0000} \textbf{-3.3}} \\ \hline
\end{tabular}
\label{tab:RM}
% }
\end{table}

In order to reduce the computational complexity of the model and enhance the inference speed, downsampling operations are applied to the width and height of keys and values in both MGF(ii) and MGF(i). As shown in Table~\ref{tab:down}, in Method 1, without downsampling, the VisEvent yields a PR of 75.8\% and SR of 59.3\%, while COESOT results in a PR of 76.1\% and SR of 67.7\%. Introducing downsampling in Method 2, with a downsampling rate of 4, leads to improvements of 0.7\% in PR and 0.8\% in SR for both datasets, compared to Method 1. Additionally, Method 2 reduces the computational cost by 0.567 $G$ and speeds up inference by 4.31 $fps$, compared to Method 1. Compared to Method 3, downsampling by a factor of 17, Method 2 demonstrates superior performance and faster inference speed on both datasets. The results from these three experiments conclusively demonstrate that downsampling the width and the height of keys and values accelerates the model's inference process, leading to a notable enhancement in performance.

\begin{table*}[!t]
\caption{{An ablation study of the effect of the Downsampling rates in the Mutually-Guided Fusion module. ``K" represents the downsampling rates.}}\vspace{2mm}
\centering
\scriptsize
\resizebox{1.0\columnwidth}{!}{
\begin{tabular}{c|c|c|ccc|ccc|cc|cc}
\hline
\multirow{3}{*}{Models} & \multirow{3}{*}{Macs} & \multirow{3}{*}{FPS} & \multicolumn{3}{c|}{MGF(ii)}                             & \multicolumn{3}{c|}{MGF(i)}                             & \multicolumn{2}{c|}{VisEvent}                                        & \multicolumn{2}{c}{COESOT}                                           \\ \cline{4-13} 
                        &                       &                      & \multicolumn{3}{c|}{K}             & \multicolumn{3}{c|}{K}             & \multicolumn{1}{c|}{\multirow{2}{*}{PR}}    & \multirow{2}{*}{SR}    & \multicolumn{1}{c|}{\multirow{2}{*}{PR}}    & \multirow{2}{*}{SR}    \\ \cline{4-9}
                        &                       &                      & \multicolumn{1}{c|}{w/o} & \multicolumn{1}{c|}{4} & 17 & \multicolumn{1}{c|}{w/o} & \multicolumn{1}{c|}{4} & 17 & \multicolumn{1}{c|}{}                       &                        & \multicolumn{1}{c|}{}                       &                        \\ \hline
1                       & 35.724 G               & 39.82 fps            & \multicolumn{1}{c|}{\checkmark}    & \multicolumn{1}{c|}{}  &   & \multicolumn{1}{c|}{\checkmark}    & \multicolumn{1}{c|}{}  &   & \multicolumn{1}{c|}{75.8}                   & 59.3                   & \multicolumn{1}{c|}{76.1}                   & 67.7                   \\
2                       & 35.157 G               & 44.13 fps            & \multicolumn{1}{c|}{}    & \multicolumn{1}{c|}{\checkmark}  &   & \multicolumn{1}{c|}{}    & \multicolumn{1}{c|}{\checkmark}  &   & \multicolumn{1}{c|}{\textbf{76.5}} & {\textbf{60.1}} & \multicolumn{1}{c|}{\textbf{76.8}} & {\textbf{68.4}} \\
3                       & 35.017 G               & 40.43 fps             & \multicolumn{1}{c|}{}    & \multicolumn{1}{c|}{}  & \checkmark  & \multicolumn{1}{c|}{}    & \multicolumn{1}{c|}{}  & \checkmark   & \multicolumn{1}{c|}{75.0}                   & 58.4                   & \multicolumn{1}{c|}{76.3}                   & 68.0                   \\ \hline
\end{tabular}
\label{tab:down}
}
\end{table*}

\textbf{The impact  of modelling the relation between the fused template and the search region.} 
We validate the  effectiveness of the Relation Modelling Block by removing it. The results obtained are presented in Table~\ref{tab:RM}. 
From the comparison between Method 1 and Method 2, it can be seen that the absence of the Relation Modelling Block causes a significant decrease in the performance of the model. 
These results indicate that the Relation Modelling Block plays a significant role in integrating the object information conveyed by the fused template  into the fused search region.

\begin{figure*}
\begin{center}
\includegraphics[width=1.0\linewidth]{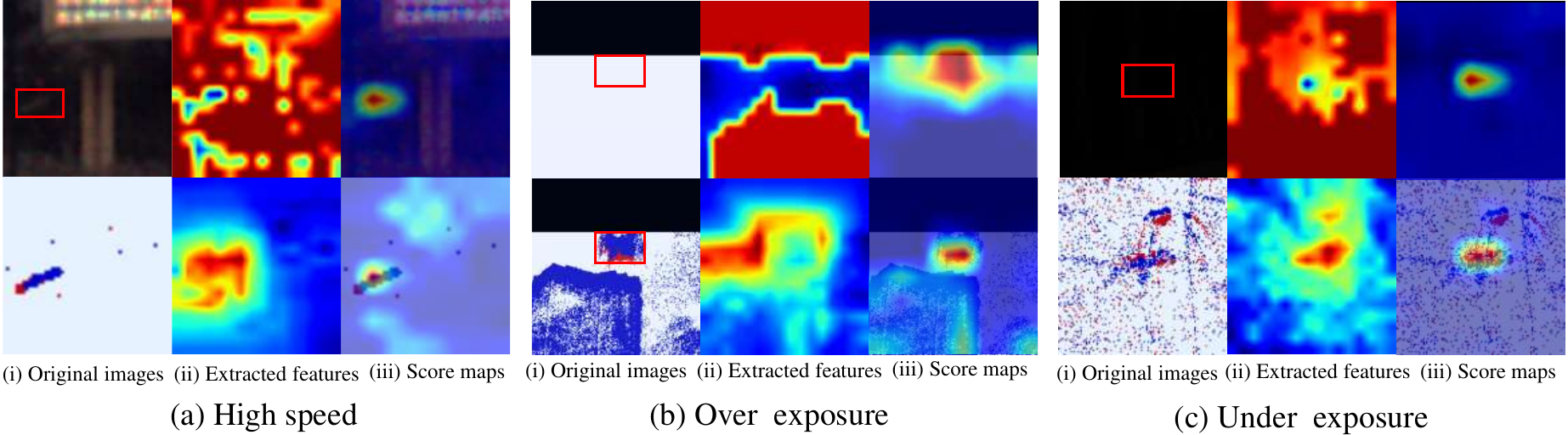}
\end{center}
\caption{Visualisation of the features and score maps. The first column represents the input images to the two modalities. The red box indicates the target. The second column shows the features extracted through the heterogeneous backbones of the two modalities. The third column presents the score maps for the two modalities after a mutual enhancement by our MGF.}
\label{fig:vis}
\end{figure*}

\textbf{Visualization.} To validate the effectiveness of our fusion module, we visualize several representative feature maps and score maps. 
For the fast-moving object in Figure~\ref{fig:vis}(a) is scattered in the RGB feature map and inaccurate in the event feature map. 
After their mutual enhancement, both modalities succeed in locating the object accurately.
For the objects in Figure~\ref{fig:vis}(b) and (c) in the Over/Under exposure scenes, the objects are almost invisible in the RGB images, but are relatively distinct in the event images. 
After being guided by the event features, the invisible objects are noted. As a result, our MGF promotes a more accurate and robust object localisation. 

\begin{figure*}[t]
\centering
\includegraphics[width=0.95\linewidth]{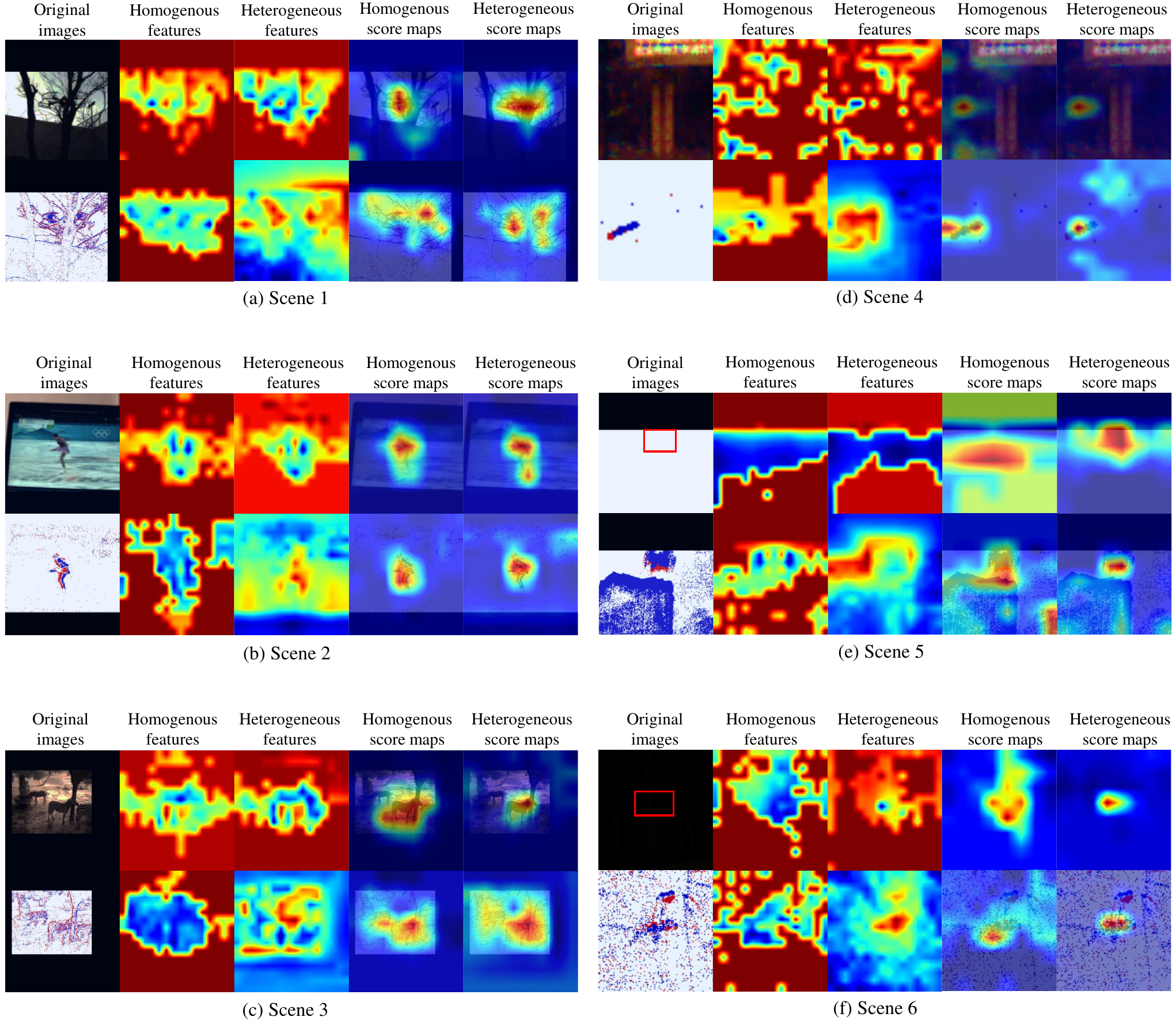}
%,height=4.5cm
\caption{Visualization of the features and score maps. The first column represents the input images to the two modalities. The red box indicates the target. The second column shows the features extracted by the homogeneous backbones (OSTrack~\cite{ostrack}). The third column presents the features extracted by the heterogeneous backbones (OSTrack~\cite{ostrack} and our Pooler) of the two modalities. The fourth column displays the score maps produced by  the two modalities, after their mutual enhancement accomplished by our MGF in experiments with the homogeneous backbones. The fifth column represents the score maps produced by the two modalities after their mutual enhancement by our MGF in experiments with the heterogeneous backbones.}
\label{fig:supplementary_vis}
\end{figure*}

 Finally, we substitute the Pooler module in the event image modality by a backbone network homologous to the RGB branch, and visualise the corresponding feature maps and score maps.
 The second column of Figure~\ref{fig:supplementary_vis} shows the features of both modalities  extracted using the same kind of backbone, specifically OSTrack~\cite{ostrack}. 
 The features of the two modalities  extracted using heterogeneous backbones are shown in the third column. 
 Specifically, the event features are extracted by our Pooler, while the RGB features are extracted by OSTrack~\cite{ostrack}. 
 The event features extracted by the RGB backbone exhibit limited efficacy in distinguishing the object region. 
 In contrast, the features derived from the Pooler are distinctly clear and effectively accentuate the object area. 
 In spite of their mutual enhancement by our MGF module, the highlighted portions fail to achieve seamless alignment with the objects in the homogeneous score maps. 
 In contrast, in the heterogeneous score maps, the highlighted part locates the object precisely.

% \begin{figure*}[t]
% \begin{center}
% \includegraphics[width=0.9\linewidth]{17_challenges.pdf}
% \caption{Precision and success plots of trackers on 17 challenges.} 
% \label{fig:challenges}
% \end{center}
% \end{figure*}

\section{Conclusion}
In this paper, we propose an end-to-end RGB-E single object tracking network. 
Our method is composed of two key components: An Event backbone (Pooler) performing Multi-Scale Pooling and a Mutually-Guided Fusion (MGF) module. 
The innovative Pooler excels in event feature extraction by leveraging the intrinsic characteristics of the event modality. 
The proposed  MGF module capitalises on the synergies between the modalities by enriching one with insights from the other. 
Thorough ablation validation conclusively demonstrates the effectiveness of our Pooler and  MGF. 
Our approach surpasses the state-of-the-art performance both on the Visevent and COESOT datasets. 
The proposed TENet is the first work taking into  considerations the sparseness property of the event modality, as well as  the real-time tracking requirements.

\section*{Acknowledgements}
This work is supported in part by the National Key Research and Development Program of China (2023YFF1105102, 2023YFF1105105), the National Natural Science Foundation of China (Grant NO. 62020106012, 62332008, 62106089, U1836218, 62336004), the 111 Project of Ministry of Education of China (Grant No.B12018), and the UK EPSRC (EP/N007743/1,MURI/EPSRC\\
/DSTL, EP/R018456/1).

\bibliographystyle{elsarticle-num}
\bibliography{references}

\end{document}